\documentclass[sigconf]{acmart}
\AtBeginDocument{%
  \providecommand\BibTeX{{%
    \normalfont B\kern-0.5em{\scshape i\kern-0.25em b}\kern-0.8em\TeX}}}

\setcopyright{acmcopyright}
\acmConference[SIGIR Resource Track]{SIGIR Resource Track}{2021}{Virtual}
\copyrightyear{2021}
\acmYear{2021}
\acmDOI{10.1145/1122445.1122456}

\usepackage{multirow}
\usepackage{hyperref}

\begin{document}
\title{WIT: Wikipedia-based Image Text Dataset for Multimodal Multilingual Machine Learning}

\author{Krishna Srinivasan}
\affiliation{\institution{Google}}
\email{krishnaps@google.com}
\author{Karthik Raman}
\affiliation{\institution{Google}}
\email{karthikraman@google.com}
\author{Jiecao Chen}
\affiliation{\institution{Google}}
\email{chenjiecao@google.com}
\author{Michael Bendersky}
\affiliation{\institution{Google}}
\email{bemike@google.com}
\author{Marc Najork}
\affiliation{\institution{Google}}
\email{najork@google.com}

\renewcommand{\shortauthors}{Srinivasan and Raman, et al.}
\begin{abstract}
  
The milestone improvements brought about by deep representation learning and pre-training techniques have led to large performance gains across downstream NLP, IR and Vision tasks. Multimodal modeling techniques aim to leverage large high-quality visio-linguistic datasets for learning complementary information (across image and text modalities). In this paper, we introduce the Wikipedia-based Image Text (WIT) Dataset\footnote{\url{https://github.com/google-research-datasets/wit}} to better facilitate multimodal, multilingual learning. WIT is composed of a curated set of 37.6 million entity rich image-text examples with 11.5 million unique images across 108 Wikipedia languages. Its size enables WIT to be used as a pretraining dataset for multimodal models, as we show when applied to downstream tasks such as image-text retrieval. WIT has four main and unique advantages. First, WIT is the largest multimodal dataset by the number of image-text examples by 3x (at the time of writing). Second, WIT is massively multilingual (first of its kind) with coverage over 100+ languages (each of which has at least 12K examples) and provides cross-lingual texts for many images. Third, WIT represents a more diverse set of concepts and real world entities relative to what previous datasets cover. Lastly, WIT provides a very challenging real-world test set, as we empirically illustrate using an image-text retrieval task as an example.

\end{abstract}
\maketitle

\section{Introduction}

Deep learning has fundamentally revolutionized the fields of NLP, IR and Vision via our ability to have a rich semantic understanding of texts and images. 
Notable examples of this include Deep CNN models \cite{simonyan2014very, szegedy2016rethinking} which set the bar for standard vision tasks like image recognition and image classification.
Attention based transformer models \cite{vaswani2017attention} like BERT \cite{devlin2018bert} have likewise enabled achieving new benchmark performance across a myriad of text understanding / NLP / IR tasks.
These transformational advances have also found their way to multimodal tasks such as image-text retrieval / search \cite{hodosh2013framing} and image captioning \cite{vinyals2015show, zhou2020unified}. 
Multimodal models -- such as ViLBERT \cite{lu2019vilbert}, UNITER \cite{chen2019uniter}, Unicoder-VL \cite{li2019unicoder} amongst others \cite{li2019visualbert, su2019vl, alberti2019fusion}  -- are able to jointly model the complex relationships between text and visual inputs leading to wins in downstream tasks like image search, Visual Question Answering (VQA) \cite{antol2015vqa} and Visual Commonsense Reasoning (VCR) \cite{zellers2019recognition}.

\begin{figure}[ht]
  \centering
  \includegraphics[width=\linewidth]{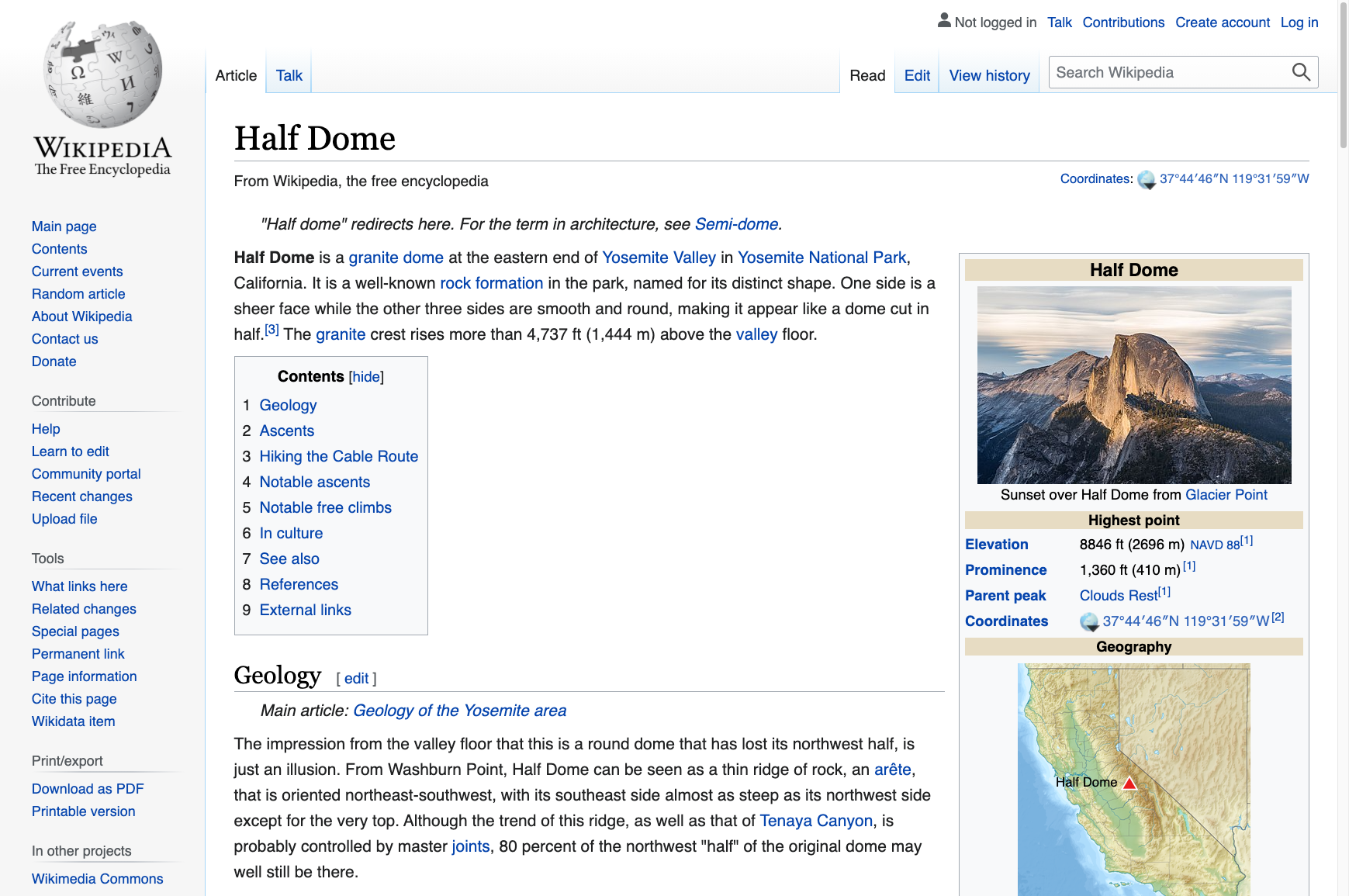}
  \caption{The Wikipedia page for Half Dome, Yosemite, California via Wikimedia Commons. \label{fig:half-dome-wiki}}
  \Description{Wikipedia page for Half Dome, Yosemite, California. (\url{https://en.wikipedia.org/Half_Dome}).}
\end{figure}


Accompanying the modeling improvements across these advancements, an equally critical aspect is the leveraging of massive datasets to enrich representation learning -- often via unsupervised \emph{pretraining}.
Increasingly, the efficacy of a model correlates strongly with the size and quality of pretraining data used.
For instance, cutting-edge language models like BERT \cite{devlin2018bert} and T5 \cite{t5paper} rely on increasingly larger text datasets spanning from those in the O(100M) range like Wikipedia, BooksCorpus \cite{zhu2015aligning} to datasets with billions of examples like C4 \cite{t5paper} and mC4 \cite{xue2020mt5}.
Similarly, vision models \cite{dosovitskiy2020image} are reliant on large corpora, such as ImageNet-21k \cite{deng2009imagenet} -- which with 14M images is among the largest public datasets.
This scale is important since studies have shown performance increases logarithmically with dataset size \cite{sun2017revisiting}.
Another key dimension of language datasets is the number of languages covered.
By transitioning from English-only to highly multilingual language datasets, models like mT5 \cite{xue2020mt5} and mBERT \cite{wu2019beto}, are an important step for researchers driving globally, equitable availability of information.


Multimodal visio-linguistic models are no different, and rely on a rich dataset to help them learn to model the relationship between images and texts. 
However as seen in Table~\ref{tab:image-text-dataset}, the scale of current public datasets pales in comparison to image-only or text-only ones, with the 30K-sized Flickr \cite{young2014image} and 3.3M-sized Conceptual Captions (CC) \cite{sharma2018conceptual} being among the largest ones.
Having large image-text datasets can significantly improve performance, as a couple of recent works \cite{radford2021learning, jia2021scaling} have shown by leveraging larger noisy (proprietary) datasets.
Furthermore the lack of language coverage in these existing datasets (which are mostly only in English) also impedes research in the multilingual multimodal space -- which we consider a lost opportunity given the potential shown in leveraging images (as a language-agnostic medium) to help improve our multilingual textual understanding \cite{singhal2019learning} or even translate \cite{hewitt2018learning}.

\begin{figure}[ht]
  \centering
  \includegraphics[width=\linewidth]{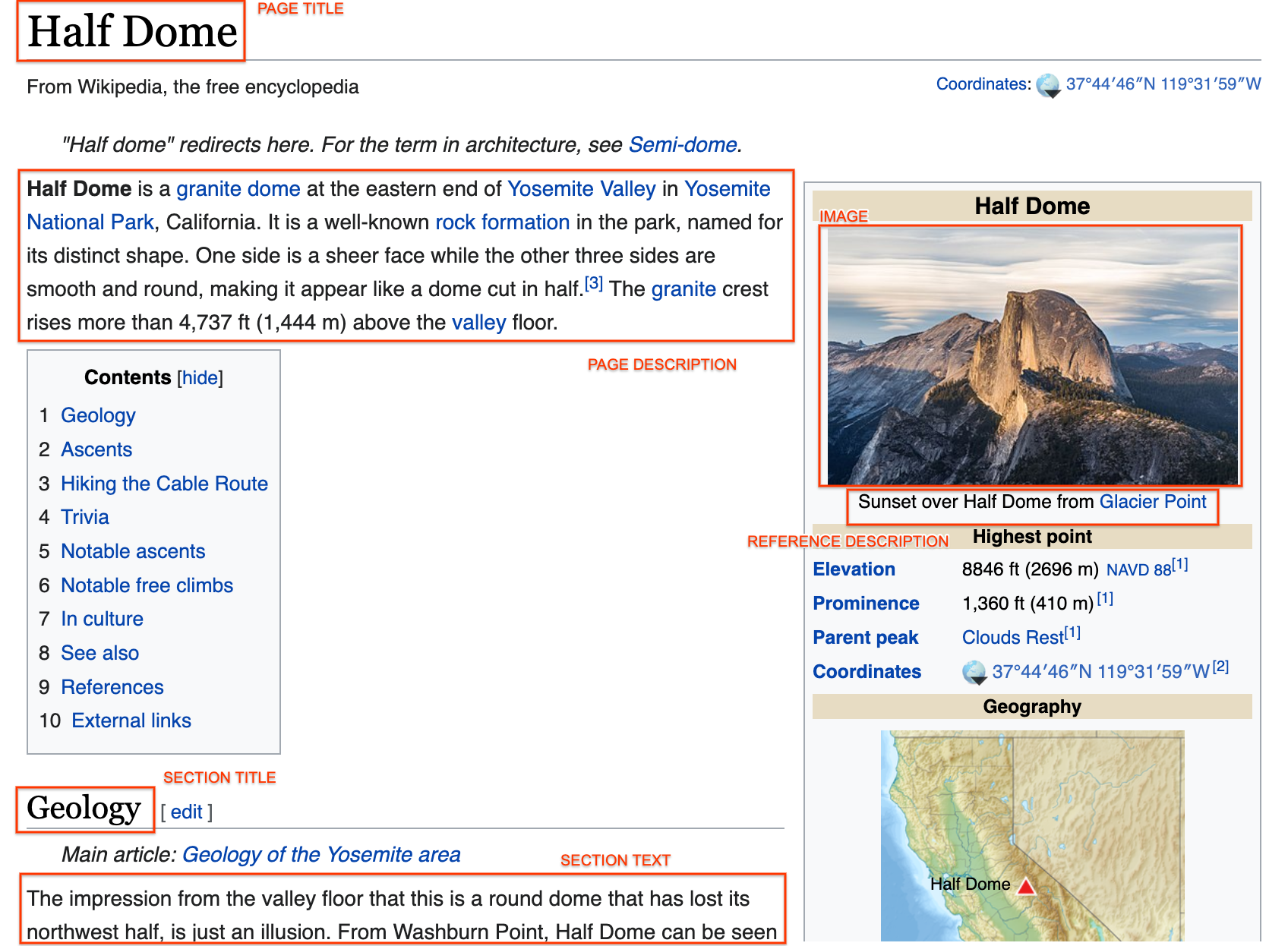}
  \caption{The Wikipedia page for Half Dome, Yosemite, California via Wikimedia Commons with examples of the different fields extracted and provided in WIT. \label{fig:half-dome-example} }
  \Description{Wikipedia page for Half Dome, Yosemite, California. (\url{https://en.wikipedia.org/Half_Dome}).}
\end{figure}

To address these challenges and advance research on multilingual, multimodal learning we present the  Wikipedia-based Image Text (WIT) Dataset.
WIT is created by extracting multiple different texts associated with an image (\emph{e.g.,} the reference description seen in Fig~\ref{fig:half-dome-example}) from Wikipedia articles and Wikimedia image links.
This was accompanied by rigorous filtering to only retain high quality image-text associations.
The resulting dataset contains over 37.6 million image-text sets and spans 11.5 million unique images -- making WIT the largest multimodal dataset at the time of writing.
Furthermore WIT provides unparalleled multilingual coverage -- with 12K+ examples in each of 108 languages (53 languages have 100K+ image-text pairs).

It is worth pointing out that by leveraging Wikipedia's editing, verification and correction mechanism, WIT is able to ensure a high-quality bar.
In particular, this use of a curated source like Wikipedia contrasts with the approach used to create other existing datasets (\emph{e.g.} CC \cite{sharma2018conceptual}) which rely on extracting annotations from web crawls.
We verified the curated quality of the WIT dataset via an extensive human-annotation process (nearly 4400 image-text examples and 13K judgments across 7 languages), with an overwhelming majority (98.5\%) judging the randomly sampled image-text associations favorably.

Empirical results on image-text retrieval tasks (both zero-shot {\emph i.e.,} pretrained model, as well as finetuned model evaluations) demonstrate the potency of the data.
The vast richness of Wikipedia texts and images (grounded in a diverse set of real-world entities and attributes) also means that WIT provides for a realistic evaluation set -- one that we demonstrate to be challenging for models trained using existing datasets.

\begin{table}
  \caption{Existing publicly available image-text datasets pale in comparison to text-only datasets (e.g., mC4 with O(Billions) of examples in 100+ languages) and image-only datasets (e.g., 14M in ImageNet-21k).}
  \vspace{-0.5em}
  \label{tab:image-text-dataset}
  \begin{tabular}{|c|c|c|c|}
    \hline 
    \textbf{Dataset} 
    & \textbf{Images}
    & \textbf{Text}
    & \textbf{Languages}    \\ 
    \hline 
    \text{Flickr30K \cite{young2014image}} 
    & \text{32K}
    & \text{158K}
    & \text{< 8}    \\  
    \hline 
    \text{SBU Captions \cite{ordonez2011im2text}} 
    & \text{$\sim$1M}
    & \text{$\sim$1M}
    & \text{1}    \\  
    \hline 
    \text{MS-COCO \cite{lin2014microsoft}} 
    & \text{$\sim$330K}
    & \text{$\sim$1.5M}
    & \text{< 4}    \\  
    \hline 
    \text{CC \cite{changpinyo2021conceptual}} 
    & \text{$\sim$3.3M}
    & \text{$\sim$3.3M}
    & \text{1}    \\  
    \hline 
    \textbf{WIT} 
    & \textbf{11.5M}
    & \textbf{37.6M}
    & \textbf{108}    \\  
  \hline
\end{tabular}
\vspace{-1em}
\end{table}
\section{Related Work}

\textbf{Visio-Linguistic (VL) datasets:}
Flickr30K \cite{young2014image} was among the first datasets that helped drive early research in this space.
Similar to other such early datasets (\emph{e.g.} the 330k example MS-COCO), it was created by having crowd sourced (Mechanical Turk) workers provide captions for $\sim$30K images (sampled from Flickr).
While the explicit human-based captioning helps ensure quality, the resulting datasets have been recognized as insufficient for significant real-world improvements given that they are small and expensive to construct \cite{elliott2017findings, zhou2018visual}.
Furthermore, this manual effort has meant extending to other languages has proven to be quite challenging.
Consequently there exists only a handful of non-English data collections such as Multi30K-DE (German) \cite{elliott2016multi30k}, DeCOCO  (German) \cite{hitschler2016multimodal}, Multi30K-FR (French) \cite{elliott2017findings}, Multi30K-CS (Czech) \cite{barrault2018findings}, YJCaptions26k (Japanese) \cite{yj26kdataset} and MS-COCO-CN (Chinese) \cite{li2019coco}.

An alternative paradigm to creating such datasets is demonstrated by the Conceptual Captions (CC) dataset \cite{sharma2018conceptual}.
By leveraging the alt-text annotations for images from a web crawl, the resulting dataset was significantly larger than previous ones ($\sim$3.3M image-text pairs). 
The drawback with this approach is the reliance on complex filtering rules and systems to ensure data quality.
Unfortunately this makes these \emph{extraction}-based datasets -- like CC and the recently proposed CC12M \cite{changpinyo2021conceptual} -- hard to extend and significantly impacts their coverage and diversity.
Perhaps unsurprisingly, the complex filtering logic has meant that this approach has so far only been successfully applied to curate English data collections.

WIT looks to achieve the best of both worlds by leveraging an extractive approach on a clean, curated multilingual repository of human knowledge with its accompanying images, illustrations and detailed text descriptions (Wikipedia).

\textbf{VL models:}
A slew of models have been proposed to leverage the above datasets (either for unsupervised pretraining or finetuning). 
For example, ViLBERT \cite{lu2019vilbert} uses MS-COCO and CC for pretraining a multimodal transformer based model. 
UNITER \cite{chen2019uniter} leverages these datasets and pretrains on tasks like image-text matching and word region alignment.
Similarly, models like VL-BERT \cite{su2019vl}, VisualBERT \cite{li2019visualbert}, ImageBERT \cite{qi2020imagebert}, B2T2 \cite{alberti2019fusion} and Unicoder-VL \cite{li2020unicoder}, all pretrain on CC or similar datasets using a variety of objectives and tasks.
Efficacy of these models is often studied on downstream tasks like image-text retrieval, referring expressions, image captioning, {\emph etc} using Flickr30K, MS-COCO and similar curated collections.
These models have also shown that a larger and more varied data collection, results in better performance across the board in downstream tasks. 

\section{WIT: Wikipedia Image Text Dataset}

\begin{table}[]
  \caption{Example of texts extracted for Half Dome example}
  \label{tab:wit-text-half-dome-example}
  \vspace{-0.8em}
  \begin{tabular}{|l|p{.55\linewidth}|}
    \hline \textbf{Field Name} & \textbf{Text}\\
    \hline Page Title & Half Dome, Yosemite\\
    \hline Canonical Page URL & \href{https://en.wikipedia.org/wiki/Half_Dome}{en.wikipedia.org/wiki/Half\_Dome}\\
  \hline Page Description & Half Dome is a granite dome at the eastern end of Yosemite Valley in Yosemite National Park, California. It is a well-known rock formation ...\\
  \hline Reference Description & Sunset over Half Dome from Glacier Point\\
  \hline Attribution Description & English: Half Dome as viewed from Glacier Point, Yosemite National Park, California, United States.\\
  \hline
\end{tabular}
\vspace{-0.2em}
\end{table}

We would like to marry the benefits of curated datasets like Flickr30K and MS-COCO (consistent, high quality image text pairs) with those of extractive datasets like CC (automatically created and scalable), while also creating a multilingual and heterogeneous dataset.
To do so, we leverage Wikipedia, which inherently uses crowd-sourcing in the data creation process -- via its editorial review process -- to ensure quality, freshness and accuracy of content.
However, even Wikipedia extractions cannot be directly used as is, due to a plethora of low-information (e.g., generic) image-text associations which would not help VL learning.
In the remainder of this section, we describe the WIT creation process and detail the filtering processes we introduced to ensure that only the most useful data is selected. 

\subsection{Wikipedia Crawl Data}

We started with all Wikipedia content pages (\emph{i.e.,} ignoring other pages that have discussions, comments and such).
These number about $\sim$124M pages across 279 languages.  We used a Flume \cite{chambers2010flumejava} pipeline to programatically process, filter, clean and store the Wikipedia data.
We next extracted images and different texts related to the image along with some contextual metadata (such as the page URL, the page title, description \dots).
This yielded about $\sim$150M tuples of \emph{(image data, texts data, contextual data)}, which were the input to the different filters described in the subsequent sections.
Note that there tends to be a wide variance of HTML formatting / layouts used for image captions across (and sometimes even within) Wikipedias in different languages, and hence our extraction rules needed to be particularly robust to ensure high coverage.

\begin{table}
  \caption{Statistics of the final WIT dataset and availability of different fields. Tuple refers to one entry in the dataset comprising the image, the three different possible texts and the context. Context texts include the page and (hierarchical) section titles and their respective descriptions}
  \label{tab:wit-dataset-numbers}
  \vspace{-1em}
  \begin{tabular}{|c|c|c|c|c|}
    \hline 
    \textbf{Type} 
    & Train
    & Val
    & Test
    & \textbf{Total / Unique} \\ 
    \hline 
    \textbf{Rows / Tuples} 
    & \text{37.13M}
    & \text{261.8K} 
    & \text{210.7K}
    & \textbf{37.6M}
    \\  
    \hline 
    \textbf{Unique Images} 
    & \text{11.4M}
    & \text{58K} 
    & \text{57K}
    & \textbf{11.5M}
    \\
    \hline 
    \textbf{Ref. Text} 
    & \text{16.9M}
    & \text{150K} 
    & \text{104K}
    & \textbf{17.2M / 16.7M}
    \\
    \hline 
    \textbf{Attr. Text} 
    & \text{34.8M}
    & \text{193K} 
    & \text{200K}
    & \textbf{35.2M / 10.9M}
    \\
    \hline 
    \textbf{Alt Text} 
    & \text{5.3M}
    & \text{29K} 
    & \text{29K}
    & \textbf{5.4M / 5.3M}
    \\
    \hline 
    \textbf{Context Texts} 
    & \text{-}
    & \text{-} 
    & \text{-}
    & \textbf{119.8M}
    \\
  \hline
\end{tabular}
\vspace{-1.5em}
\end{table}

\subsection{The Texts used in WIT}
The texts describing the images come from multiple different sources. The three \emph{directly} associated with the image are:

\begin{enumerate}
    \item \textbf{Reference description} (abbreviated as \emph{ref}): This is the caption that is visible on the wiki page directly below the image. This is the least common among the three (present in $\sim$24M of the tuples) but tends to be the most topical and relevant.

    \item \textbf{Attribution description} (abbreviated as \emph{attr}): This is the text found on the Wikimedia page of the image. This text is common to all occurrences of that image across all Wikipedias and thus can be in a language different to the original page article. Often this text is multilingual \emph{i.e.,} with image descriptions in multiple languages. 138M+ of the 150M tuples have this field -- though the vast majority of these are uninformative or noisy. However the remaining have rich semantic descriptions of the images that we would like to extract.

    \item \textbf{Alt-text description} (abbreviated as \emph{alt}): This is the “alt” text associated with the image. While not visible in general, it is commonly used for accessibility / screen readers. Despite this (surprisingly) we discovered that this was the least informative of the three texts and in most cases was simply the image file name (We found that of the 121M+ tuples containing this text, only a small fraction to be meaningful descriptions of the image).
\end{enumerate}

In addition to these, we also note that the context part of the tuple contains additional texts indirectly associated with the image (such as the section text or page title).
A complete example of these texts, along with other metadata fields (as illustrated in Table \ref{tab:wit-full-example}) we provide and more detailed statistics are available on the WIT dataset Github page.

\vspace{-0.5em}
\subsection{Text-based Filtering}
\label{sec:filtering-start}

To clean the low-information texts, we:
\begin{enumerate}
    \item Only retained texts that were at least of length 3.
    \item Removed any alt-text containing generic phrases such as '.png', '.jpg', ‘icon’ or ‘stub’ and also phrases with “refer to”, “alt text” .. etc.
    \item For attributions and alt-text we enforced that
    \begin{itemize}
      \item Image is either JPG or PNG (since these texts for other image types were almost always unhelpful). GIF images with a reference description were retained.
      \item For tuples without a reference description, we enforced the image is not found in the last sections of the page (\emph{i.e.,} the bibliography, external links and such).
    \end{itemize}
\end{enumerate}

\vspace{-0.5em}
\subsection{Image \& Image-Text based Filtering}

We applied the following filters on the images in the tuples:

\begin{enumerate}
    \item To ensure rich images, we required that image height and width were at least 100 pixels.
    
    \item Based on a detailed analysis, we eliminated images which were either generic or didn't have meaningful text associations. For example, images of maps are prevalent on Wikipedia for denoting locations. However since these are generic and not specific to the actual location, the text association is often incorrect and hence we removed them. Other such noise patterns included common images (\emph{e.g.,} tiny icons), placeholder images and generic \emph{missing images}.

    \item We only retained images that have a research-permissive license such as Creative Commons (the text of Wikipedia is licensed under a CC-BY-SA license).

    \item Lastly we found that certain image-text pairs occurred very frequently. These were often generic images that did not have much to do with the main article page. Common examples included flags, logos, maps, insignia and such. To prevent biasing the data, we heavily under-sampled all such images. 
    
\end{enumerate}

\vspace{-0.5em}
\subsection{Additional Filtering}

To ensure a high-quality dataset free of inappropriate content, we removed tuples with questionable images or texts as done by previous works \cite{sharma2018conceptual}.
In particular we aimed to remove pornographic / profane / violent / \dots content using multiple techniques based on sophisticated image understanding and multilingual text understanding models.
Overall these filters help improve data quality while only eliminating < 0.2\% of all tuples.

Akin to other multilingual datasets (\emph{e.g.,} mC4 \cite{xue2020mt5}), we restricted our initial version to only the top ~100 languages and hence only retained tuples for languages with 12K+ tuples.
Lastly we created partitioned the data into training, validation and test splits (with ~50K images for the latter two) by ensuring that each image only occurs in a single split.

\subsection{Analyzing the WIT Data}
As seen in Table \ref{tab:image-text-dataset}, the resulting dataset is significantly larger than previous ones with over 37M (image, text(s), context) tuples, spanning 108 languages and covering 11.5 million unique images.
Among its many unique aspects and firsts:

\begin{table}
  \caption{WIT: Image-Text Stats by Language}
  \vspace{-0.5em}
  \label{tab:wit-image-text-stats}
  \begin{tabular}{|c|c|c|c|}
    \hline 
    \textbf{Image-Text} 
    & \textbf{\# Lang}
    & \textbf{Uniq. Images}
    & \textbf{\# Lang} \\ 
  \hline
total > 1M	& 9 & images > 1M & 6 \\
  \hline
total > 500K & 10 & images > 500K & 12 \\
  \hline
total > 100K & 36 & images > 100K & 35 \\
  \hline
total > 50K	& 15 & images > 50K & 17 \\
  \hline
total > 14K	& 38 & images > 13K & 38 \\
    \hline
\end{tabular}
\vspace{-1em}
\end{table}

\begin{itemize}
    \item \textbf{Multiple texts per image}: WIT provides for multiple different kinds of texts per image.
    More than half of the tuples (19.4M) have two or more of reference, attribution and alt-texts.
    Table~\ref{tab:wit-dataset-numbers} provides some more detailed statistics of the coverage of the different texts.
    Overall with nearly 32M unique image-text pairs, WIT is nearly an order of magnitude larger than prior datasets.
    \item \textbf{Highly multilingual}: As seen in Table~\ref{tab:wit-image-text-stats}, WIT has broad multilingual coverage.
    Nearly half of the 100+ languages, contain 100K+ unique image-text tuples and 100K+ unique images.
    \item \textbf{Large cross-lingual coverage}: Images have shown great promise in helping build cross-lingual models \cite{specia2016shared, singhal2019learning}.
    WIT can be used to generate 50M+ cross-lingual pairs (\emph{i.e.,} text descriptions in different languages for the same image) from 3.1M different images using just the reference and alt texts.
    We expect this number to be even higher when counting attributes, many of which are inherently multilingual.
    \item \textbf{Contextual understanding}: WIT is also the first dataset, providing for understanding image captions \emph{in the context of} the page and surrounding text (incl. $\sim 120M$ contextual texts). For the sake of brevity we explore this in future work.
\end{itemize}

\begin{figure}[htbp]
  \centering
  \includegraphics[width=\linewidth]{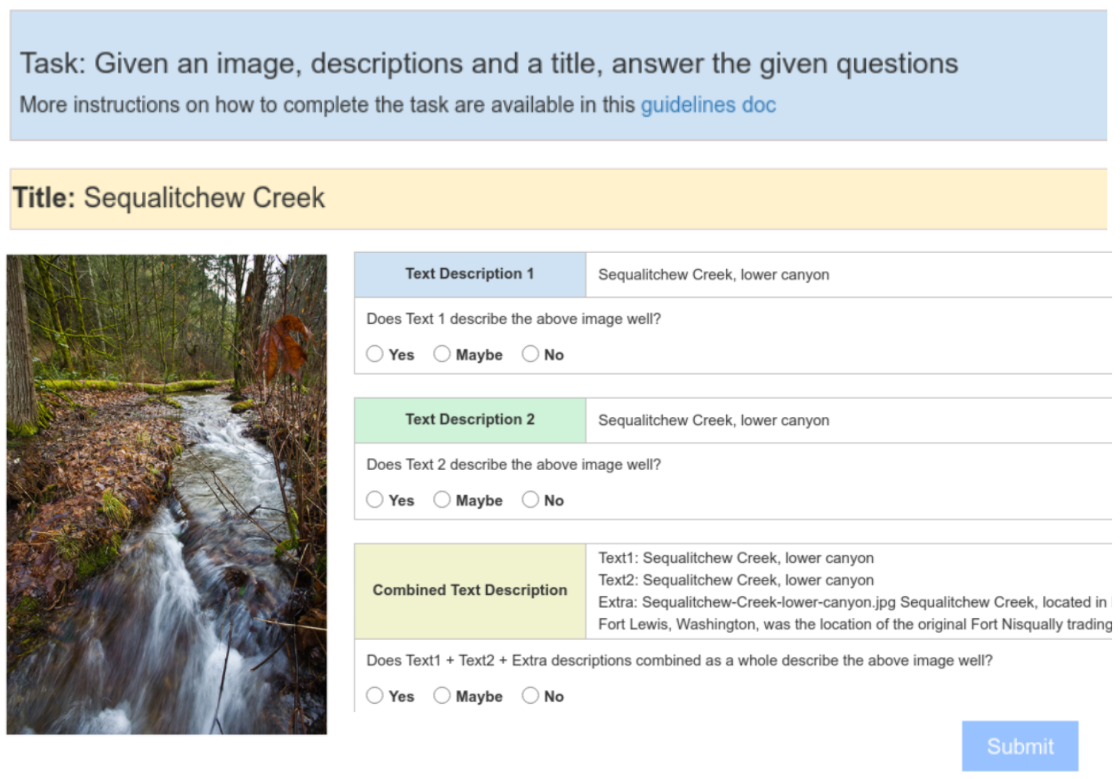}
  \vspace{-2em}
  \caption{Human Annotation Template Example \label{fig:eval-template-example}}
  \vspace{-0.8em}
  \Description{WIT Data Evaluation Template Example}
\end{figure}

\subsection{Human Annotator Validation}

To further verify the quality of the WIT dataset we performed a study using (crowd-sourced) human annotators.
As seen in Fig.~\ref{fig:eval-template-example}, we asked raters to answer 3 questions.
Given an image and the page title, raters first evaluate the quality of the attribution description and reference description in the first two questions (order randomized).
The third question understands the \emph{contextual} quality of these text descriptions given the page description and caption.
Each response is on a 3-point scale: "Yes" if the text perfectly describes the image, "Maybe" if it is sufficiently explanatory and "No" if it is irrelevant or the image is inappropriate.

\begin{table}
  \caption{Results of the human annotations of data quality. These examples and ratings are included with the dataset.}
  \vspace{-0.8em}
  \label{tab:wit-he-results}
  \begin{tabular}{|c|c|c|c|c|c|c|}
    \hline 
    \textbf{Text} 
    & \multicolumn{3}{|c|}{EN}
    & \multicolumn{3}{|c|}{non-EN} \\
  \hline
  & \%Yes & \%Maybe & \%No & \%Yes & \%Maybe & \%No  \\
  \hline
  Reference & 92.2 & 4.4 & 3.3 & 94.1 & 2.9 & 2.9 \\
  \hline
Attribute & 92.2 & 3.3 & 4.6 & 93.1 & 0.8 & 6.2 \\
  \hline
Contextual & 98.7 & 0.7 & 0.6 & 96.6 & 1.8 & 1.6 \\
  \hline
\end{tabular}
\vspace{-1.5em}
\end{table}

We randomly sampled nearly 4.4k examples for this evaluation.
To maximize rating quality we used a language identification filter on the attribution to show raters examples in the language of their expertise.
In addition to rating $\sim 3k$ examples in English, we also rated 300 examples in German, French, Spanish, Russian, Chinese and 100 examples for Hindi.
(We chose these languages to capture different language families and different sizes -- Hindi is only $65^{th}$ in size).
Each example was rated by three raters and majority label was used (Maybe being selected if no majority).
As seen from the results in Table~\ref{tab:wit-he-results}, an overwhelming majority of examples were found to be very helpful.
Both reference and attribution were found to be high-quality (with a slight edge to reference description).
The responses to the third question (which provides the page context) also validated our hypothesis that the relevance of image captions is influenced by the context as seen by the near-perfect ratings when considering the context.
Lastly we found no major difference in performance across the different languages demonstrating the multilingual data quality.


\section{Multimodal Experiments with WIT}

In this section, we empirically demonstrate the efficacy of the WIT dataset both as a pretraining dataset as well as an evaluation set for a new image-text retrieval task.

\subsection{Experiment Details}

\begin{figure}[ht]
  \centering
  \includegraphics[scale=.3]{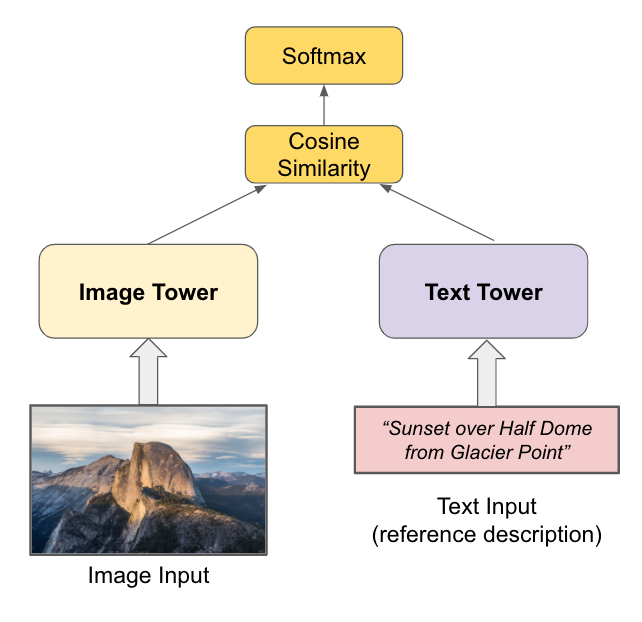}
  \Description{WIT Dual Encoder Model for Training.}
  \caption{WIT Dual Encoder Model for Training.}
  \vspace{-0.5em}
  \label{fig:wit_dual_encoder_architecture}
\end{figure}

\textbf{Model}: For this analyses, we leveraged a two-tower or \emph{dual-encoder} model, inspired by previous works that used them to learn multilingual, multimodal models \cite{singhal2019learning}.
As the name suggests, the model has two encoders -- one to encoder the text and the other to represent the images.
While the text input to the model was a bag of words, the image tower, the image was first embedded in a manner similar to \cite{juan2019graph}.
The final embeddings of these two towers is then combined using their cosine similarity, which in turn is optimized using a batch softmax loss. The dual-encoder architecture is illustrated in Figure \ref{fig:wit_dual_encoder_architecture}.
Specifically, for a batch of $n$ image-text embedding pairs, the complete $n \times n$ similarity matrix is computed (the $(i, j)$ entry being the cosine of the $i^{th}$ image embedding and $j^{th}$ text embedding) and a softmax loss applied on each of the row.
Note that only the diagonal entries are considered as positive pairs.

\textbf{Setup}:
We used a batch size of 128 for training and a batch size of 1000 for evaluation.
The learning rate was set to 5e-7
The optimizer we used was SGD with Momentum.
For the text encoder, we used a bag of words model (with ngrams of size 1 and 2).
Each ngram was mapped to an a one amongst a million vocabulary buckets using a hash-function to get a 200D embedding.
These ngram embeddings were then summed and passed through a simple FFNN and projected to a final 64D embedding, to match the size of the image encoder embedding.
The final activation function we used was ReLU.

\textbf{Evaluation}: We evaluated the models on the Flickr30K, Multi30K and MS-COCO test sets, as well as the dedicated test sets released as part of WIT.
We also spliced the WIT test sets into English-only and i18n (non-English) to understand any performance differences.
In all experiments using WIT for pretraining, we use the entire training set (\emph{i.e.,} data for all languages).
We also pretrained a model with Conceptual Captions (CC) dataset to compare against.
We used Recall@K (K = {1, 3, 5}) as the evaluation metric.

\begin{table}[]
\caption{Zero-shot evaluation for models using different text fields on WIT Image-Text Retrieval test sets}
\vspace{-0.5em}
\label{tab:wit-de-exp-retrieval-2}
\begin{tabular}{|c|c|c|c|c|c|c|c|}
\hline
\multicolumn{2}{|c|}{\textbf{Pretrain setup}} & \multicolumn{2}{c|}{\textbf{WIT-All}} & \multicolumn{2}{c|}{\textbf{WIT-EN}} & \multicolumn{2}{c|}{\textbf{WIT-I18N}} \\ \hline
\textbf{Data} & \textbf{Text} & \textbf{R@1} & \textbf{R@5} & \textbf{R@1} & \textbf{R@5} & \textbf{R@1} & \textbf{R@5}  \\ \hline
\textbf{WIT}     
& \textbf{ref}     
& \text{0.126}     
& \text{0.258}     
& \text{0.169}     
& \text{0.358}     
& \text{0.114}     
& \text{0.236} \\ \hline
\textbf{WIT}     
& \textbf{attr}     
& \text{0.293}     
& \text{0.55}     
& \text{0.272}     
& \text{0.523}     
& \text{0.293}     
& \text{0.523} \\ \hline
\textbf{WIT}     
& \textbf{ref+attr}     
& \textbf{0.346}     
& \textbf{0.642}     
& \textbf{0.344}     
& \textbf{0.64}     
& \textbf{0.344}     
& \textbf{0.633} \\ \hline
\textbf{CC}     
& \textbf{text}     
& \text{0.048}     
& \text{0.122}     
& \text{0.072}     
& \text{0.186}     
& \text{0.041}     
& \text{0.11} \\ \hline
\end{tabular}
\vspace{-0.5em}
\end{table}

\subsection{Evaluating a zero-shot pretrained model}

A common evaluation of image-text datasets is as a pretraining dataset for a model, which is then directly applied to a downstream task -- in our case image-text retrieval -- without any finetuning (\emph{i.e.,} zero-shot).
Since WIT contains multiple different texts associated with an image, we first set about understanding the effect of pretraining models on different fields.
As seen in Table~\ref{tab:wit-de-exp-retrieval-2}, the different WIT models all perform quite well on both English and non-English sets.
The strongest performance was consistently obtained by the concatenation of reference and attribution descriptions -- which we now default to for subsequent experiments.
It is worth noting that the model pretrained on CC lags behind those trained on WIT, even on the English-only test set.

To better understand this, we next evaluated the WIT and CC models (in this zero-shot manner) on popular English test collections from Flickr30K and MS-COCO which are more similar to CC.
As seen in Table \ref{tab:wit-de-exp-retrieval-1}, the multilingual WIT model trails the English CC model on these collections, though not as significantly as the gap between WIT and CC on the heldout WIT test sets.

\begin{table}[]
\caption{Zero-shot Evaluation on Flickr30K, MS-COCO and WIT test sets for Image-Text Retrieval Task}
  \vspace{-0.5em}
\label{tab:wit-de-exp-retrieval-1}
\begin{tabular}{|c|c|c|c|c|c|c|}
\hline
\multirow{2}{*}{\textbf{Pretrain}} & \multicolumn{2}{c|}{\textbf{MS-COCO}} & \multicolumn{2}{c|}{\textbf{Flickr30K}} & \multicolumn{2}{c|}{\textbf{WIT-ALL}} \\ \cline{2-7}
    & \textbf{R@1} & \textbf{R@5} & \textbf{R@1} & \textbf{R@5} & \textbf{R@1} & \textbf{R@5}  \\ \hline
%
%
\textbf{WIT-ALL}     
& \text{0.074}     
& \text{0.228}     
& \text{0.054}     
& \text{0.165}     
& \textbf{0.346}     
& \textbf{0.642}\\  \hline
\textbf{CC}     
& \textbf{0.145}     
& \textbf{0.385}     
& \textbf{0.111}     
& \textbf{0.32}     
& \text{0.048}     
& \text{0.122}\\   \hline
\end{tabular}
  \vspace{-0.1em}
\end{table}

\subsection{Understanding multilingual performance}

Since WIT encompasses examples from 100+ languages, we next evaluated how multilingual the WIT-based models are.
For this, we used Multi30K's three language test sets (Czech (CS), German (DE) and French (FR)).
We generated similar language subset datasets from the WIT test set for the same languages (CS, DE, FR) and used that for evaluation.
As shown in Table \ref{tab:wit-de-exp-retrieval-3}, both models struggle on the Multi30K dataset, though again the WIT model shines on the held-out WIT test set.
Similar to the Flickr30k dataset, the Multi30k datasets are quite different from the WIT datasets (as we discuss in Sec.~\ref{sec:discussion}) which may explain this behavior.

\begin{table}[]
\caption{Zero-shot Evaluation on Multi30K and WIT I18N test sets (CS, DE, FR) for Image-Text Retrieval Task}
\vspace{-0.5em}
\label{tab:wit-de-exp-retrieval-3}
\begin{tabular}{|c|c|c|c|c|c|c|}
\hline
\multirow{2}{*}{\textbf{Exp}} & \multicolumn{3}{c|}{\textbf{Multi30K-R@5}} & \multicolumn{3}{c|}{\textbf{WIT-R@5}} \\ \cline{2-7}
    & \textbf{CS} & \textbf{DE} & \textbf{FR} & \textbf{CS} & \textbf{DE} & \textbf{FR}  \\ \hline
%
\textbf{WIT-ALL}     
& \textbf{0.006}     
& \textbf{0.005}     
& \textbf{0.006}     
& \textbf{0.553}     
& \textbf{0.562}     
& \textbf{0.599} \\ \hline
\textbf{CC}     
& \text{0.004}     
& \textbf{0.005}   
& \text{0.004}     
& \text{0.096}     
& \text{0.084}     
& \text{0.104} \\ \hline

\end{tabular}
\vspace{-0.1em}
\end{table}

\subsection{Evaluation On (Image, Wiki Page Title) Retrieval Task}

Lastly, we evaluated on a real-world task that's based on Wikipedia.
This retrieval task requires identifying images that can be found on a given Wikipedia page, using only the page title. 
We ran this evaluation in both a zero-shot setting (\emph{i.e.,} pretrained model directly) and with finetuning on the training set.
Unlike the above experiments, here the input to the  text encoder was the page title directly.
The evaluation was done with the held-out WIT test split using the page title as text.
From Table~\ref{tab:wit-de-exp-retrieval-4}, we clearly observe a large performance gain on this task using WIT relative to the CC model both with and without finetuning.

\begin{table}[]
\caption{Zero-shot and Finetuned Evaluation on Wiki (Image, Page Title) test set for Retrieval Task}
\label{tab:wit-de-exp-retrieval-4}
\vspace{-0.5em}
\begin{tabular}{|c|c|c|c|c|}
\hline
\multirow{2}{*}{\textbf{Exp}} & \multirow{2}{*}{\textbf{Finetuning}} & \multicolumn{3}{c|}{\textbf{WIT-All}}  \\ \cline{3-5}
            &  & \textbf{R@1} & \textbf{R@3} & \textbf{R@5}  \\ \hline
%
\textbf{WIT-EN}     
& \text{None}     
& \text{0.067}     
& \text{0.122}     
& \text{0.152} \\ \hline
\textbf{CC}     
& \text{None}     
& \text{0.012}     
& \text{0.024}     
& \text{0.032} \\ \hline
\textbf{WIT-ALL}     
& \text{WIT-ALL}     
& \textbf{0.1}     
& \textbf{0.174}     
& \textbf{0.214} \\ \hline
\textbf{CC}     
& \text{CC}     
& \text{0.01}     
& \text{0.021}     
& \text{0.029} \\ \hline
\end{tabular}
\end{table}

\subsection{Discussion}
\label{sec:discussion}

The above experiments clearly demonstrated that WIT-based pretrained models perform extremely well (5x+ gains) on the evaluation sets based on Wikipedia data. However, the models do not do as well on other image-text datasets (Flickr30K/Multi30k and MS-COCO).
Since the WIT dataset is not lacking in size or diversity, we probed further into what makes these evaluation sets so different from each other.

\subsubsection{Vocabulary Analysis}

We first analyzed the vocabulary of the two datasets we used for pretraining : WIT and CC.
Since Wikipedia is entity heavy with a diverse concept pool, we suspected that the vocabulary of the WIT dataset may reflect this. 
As shown in Table \ref{tab:wit-vocab}, this was the case with over 72\% of WIT unigrams occurring 3 times or less (vs. 43\% for CC). 

\begin{table}[]
  \caption{Vocabulary Comparison}
  \label{tab:wit-vocab}
  \vspace{-0.5em}
  \begin{tabular}{|l|r|r|r|}
    \hline 
    \textbf{Dataset} 
    & \textbf{Unigrams}
    & \textbf{freq <= 3}
    & \textbf{pct freq <= 3}\\ 
    \hline 
    \text{CC} 
    & \text{149,924}
    & \text{63,800}
    & \text{42.55\%}\\  
    \hline 
    \text{WIT (ref)} 
    & \text{867,906}
    & \text{625,100}
    & \textbf{72.02\%}\\  
  \hline
\end{tabular}
\end{table}

\subsubsection{Language Model}

This difference is even more stark when compared to the test collections used for evaluation (COCO and Flickr).
When we compared the unigram distributions of different data sets using the Jensen-Shannon Divergence (JSD), we found a massive difference in the vocabularies and concept coverage of the data (see Table~\ref{tab:wit-lang-model}).
While the fact that less than a sixth of WIT is English skews these results slightly, the gap between the English-only slice and other datasets remains sizeable.

\begin{table}
  \caption{Language Model Comparison}
  \label{tab:wit-lang-model}
  \vspace{-0.5em}
  \begin{tabular}{|c|c|}
    \hline 
    \textbf{Dataset A vs B} 
    & \textbf{JSD}\\ 
    \hline 
    \text{Flickr vs Flickr Test}
    & \text{0.1679}\\  
    \hline 
    \text{COCO vs COCO Test}
    & \text{0.1008}\\  
    \hline 
    \text{CC vs Flickr Test}
    & \text{0.4844}\\  
    \hline 
    \text{CC vs COCO Test}
    & \text{0.4746}\\  
    \hline 
    \text{CC vs WIT}
    & \text{0.3825}\\  
    \hline 
    \textbf{WIT vs Flickr Test}
    & \textbf{0.6007}\\  
    \hline 
    \textbf{WIT vs COCO Test}
    & \textbf{0.5957}\\  
  \hline
\end{tabular}
\vspace{-0.1em}
\end{table}

\subsubsection{Image entity Analysis}

Part of the reason for this difference is the broad coverage of entities in the WIT dataset.
Using an image classification model to tag all WIT images with entities, we found that amongst the $\sim$4.5M entities identified, a large number ($\ge 80\%$ \emph{i.e.,} $\sim$3.68M) of the entities occur 3 times or less.
Thus similar to the texts, the image data too is very diverse with not much repetition.


\subsubsection{Key differences in texts}

Text fields in WIT often tend to be descriptive, verbose and use specific terminology.
However this causes a mismatch when evaluated on the test collections, which are often terse single line captions of common words and objects.
The choice of bag of words likely exacerbates this issue.
Perhaps the most important difference is the use of specifics vs general words.
As found in the CC work \cite{sharma2018conceptual}, text hypernymization was crucial to creating a dataset closer to those used for evaluation.
For example a text like \texttt{Two sculptures by artist Duncan McKellar adorn trees outside the derelict Norwich Union offices in Bristol, UK} would be transformed to \texttt{sculptures by person adorn trees outside the derelict offices} so as to remove specifics (person names, locations, times etc ..).
This is likely the biggest reason why our trained models underperformed on the existing collections.
While there are benefits and drawbacks of such hypernymization, we would like to add this in future versions.
However there remains significant challenges doing such replacements for a 100+ language dataset consistently and with high quality across languages.

\section{Future Work}

In our eagerness and excitement to share the WIT Dataset with the research community, we have just touched the tip of the iceberg by starting out with an image-text retrieval task using a simple dual encoder model. Given the superior performance of cross-attention multimodal transformer models, WIT can potentially be used in lieu of or in addition to the existing pretraining datasets in models as illustrated by UNITER, Unicoder-VL, VL-BERT, … etc. A range of new i18n tasks can be formulated with WIT as the basis for VQA, VCR and many others. Similarly, more specific i18n retrieval or captioning tasks for low resource languages are yet to be explored. There is also the possibility of using multimodality to enhance multilingual performance. WIT Dataset provides a crosslingual corpus of text for the same image which could aid in this idea. We also hope to leverage the knowledge base and entities and attributes of WIT to improve Q\&A tasks.

\section{Conclusion}

In this paper we introduced the Wikipedia Image Text (WIT) dataset -- the largest (at time of writing), multilingual, multimodal, context-rich dataset.
By extracting texts associated with images and their surrounding contexts from over a 100 languages, WIT provides for a rich and diverse dataset.
As a result, it is well suited for use in a myriad of ways including pretraining multimodal models, finetuning image-text retrieval models or building cross-lingual representations to name a few.
Our detailed analysis and quality evaluation, validate that WIT is a high quality dataset with strong image-text alignment.
We also empirically demonstrated the use of this dataset as both a pretraining and finetuning set, and in the process uncovered some shortcomings of existing datasets.
We believe this can serve as a rich resource to drive research in the multilingual, multimodal space for years to come and enable the community to building better and more robust visio-linguistic models well suited to real world tasks.

\begin{acks}
We thank Beer Changpinyo, Corinna Cortes, Joshua Gang, Chao Jia, Ashwin Kakarla, Mohammad Khan, Mike Lee, Zhen Li, Piyush Sharma, Radu Soricut, Yunhsuan Sung, Ashish Vaswani, Yinfei Yang, and many others for their insightful feedback and help.
\end{acks}

\bibliographystyle{ACM-Reference-Format}
\bibliography{wit}

\appendix
\section{Appendix}

\begin{figure*}[htb]
  \centering
  \includegraphics[width=\linewidth]{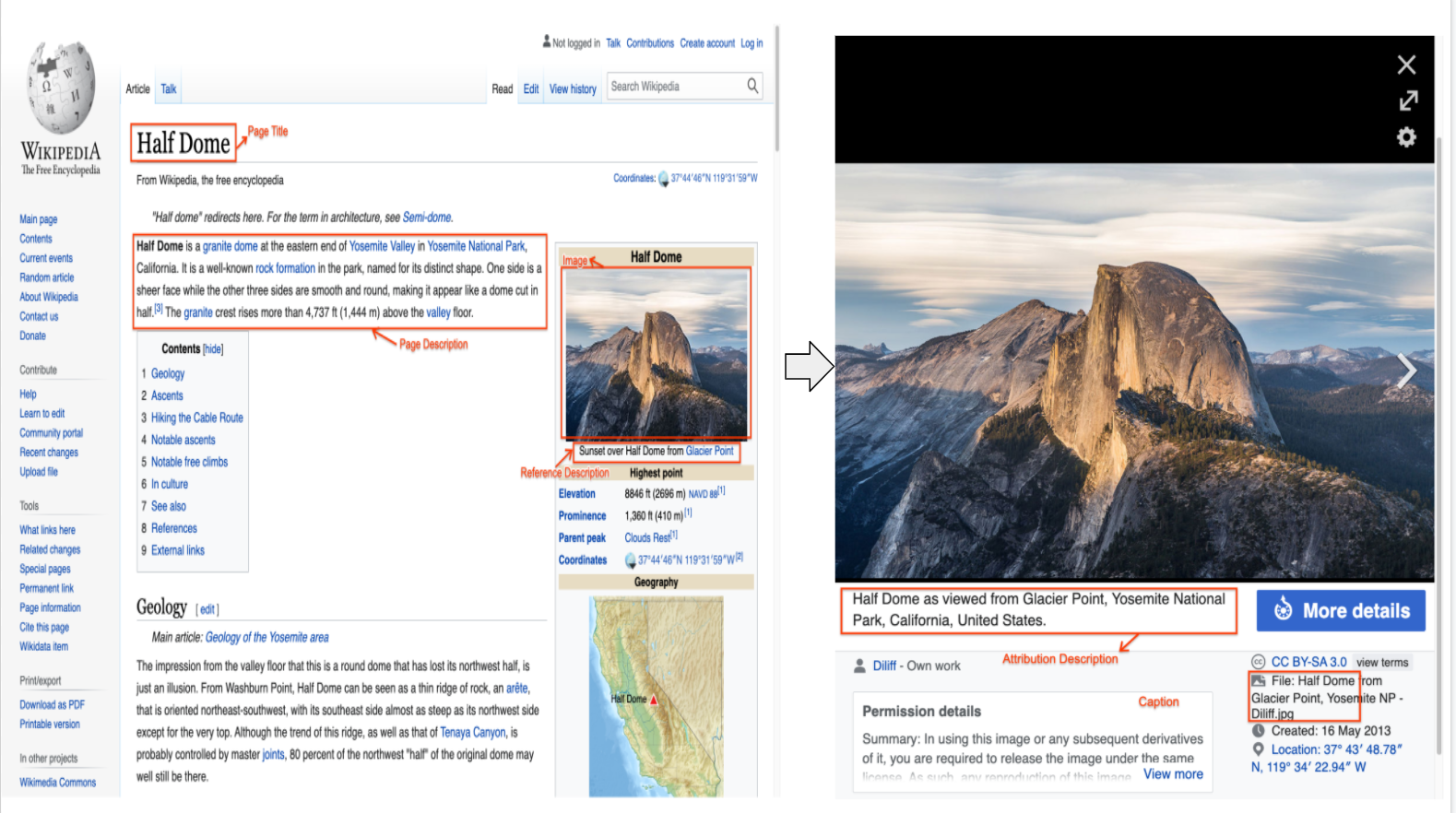}
  \caption{WIT Image-Text Example with All Text Annotations}
  \Description{WIT Image-Text Example with All Text Annotations.
  (\url{https://en.wikipedia.org/wiki/Half_Dome}).}
  \label{fig:example-with-annotation}
\end{figure*}

\begin{table*}
  \caption{WIT: A Full Example Illustration : Half Dome, Yosemite}
  \label{tab:wit-full-example}
  \begin{tabular}{|p{.40\linewidth}|p{.10\linewidth}|p{.45\linewidth}|}
    \hline
    \textbf{FIELD NAME}
    & \textbf{FIELD TYPE}
    & \textbf{VALUE}\\

    \hline
    canonical\_page\_url
    & string
    & https://en.wikipedia.org/wiki/Half\_Dome\\

    \hline
    image\_url
    & string
    & 
    \href{http://upload.wikimedia.org/wikipedia/commons/d/d6/Half_Dome_from_Glacier_Point\%2C_Yosemite_NP_\-_Diliff.jpg}{Link to Yosemite JPEG media}\\
    
    \hline
    page\_title
    & string
    & Half Dome\\

    \hline
    page\_description
    & string
    & Half Dome is a granite dome at the eastern end of Yosemite Valley in Yosemite National Park, California. It is a well-known rock formation in the park, named for its distinct shape. One side is a sheer face while the other three sides are smooth and round, making it appear like a dome cut in half. The granite crest rises more than 4,737 ft above the valley floor.\\

    \hline
    section\_text
    & string
    & Half Dome is a granite dome at the eastern end of Yosemite Valley in Yosemite National Park, California. It is a well-known rock formation in the park, named for its distinct shape. One side is a sheer face while the other three sides are smooth and round, making it appear like a dome cut in half. The granite crest rises more than 4,737 ft (1,444 m) above the valley floor.\\

    \hline
    language
    & string
    & en\\

    \hline
    is\_main\_image
    & bool
    & TRUE\\

    \hline
    mime\_type
    & string
    & image/jpeg\\

    \hline
    original\_height
    & int
    & 2988\\

    \hline
    original\_width
    & int
    & 4752\\

    \hline
    filtered\_reference\_description
    & string
    & Sunset over Half Dome from Glacier Point\\

    \hline
    filtered\_caption
    & string
    & empty value\\

    \hline
    filtered\_attribution\_description
    & string
    & English: Half Dome as viewed from Glacier Point, Yosemite National Park, California, United States.\\

    \hline
    attribution\_passes\_lang\_id
    & bool
    & TRUE\\

    \hline
    page\_changed\_recently
    & bool
    & TRUE\\

    \hline
    section\_title
    & string
    & empty value\\

    \hline
    hierarchical\_section\_title
    & string
    & Half Dome\\

    \hline
    page\_url
    & string
    & http://en.wikipedia.org/wiki/Half\_Dome\\

  \hline
\end{tabular}
\end{table*}


\begin{table*}
  \caption{Language Stats}
  \label{tab:wit-lang-stats-1}
  \begin{tabular}{|c|c|c|c|c|c|c|}
    \hline
    \textbf{Lang}
    & \textbf{Dataset}
    & \textbf{\# Examples}
    & \textbf{\# Uniq. Img}
    & \textbf{\# Ref}
    & \textbf{\# Attr}
    & \textbf{\# Cap}
    \\
  \hline

af & val & 603 & 424 & 371 & 467 & 86 \\
\hline
af & train & 93,755 & 85,728 & 47,173 & 85,180 & 18,403 \\
\hline
af & test & 542 & 497 & 276 & 512 & 116 \\
\hline
an & val & 158 & 127 & 72 & 124 & 43 \\
\hline
an & train & 27,134 & 24,904 & 9,317 & 24,900 & 7,584 \\
\hline
an & test & 186 & 172 & 73 & 178 & 47 \\
\hline
ar & val & 3,888 & 2,630 & 1,520 & 3,173 & 98 \\
\hline
ar & train & 616,618 & 534,440 & 172,069 & 590,016 & 22,228 \\
\hline
ar & test & 3,595 & 3,015 & 1,065 & 3,487 & 134 \\
\hline
arz & val & 1,288 & 1,188 & 68 & 1,262 & 7 \\
\hline
arz & train & 247,771 & 243,215 & 9,387 & 245,931 & 1,700 \\
\hline
arz & test & 1,376 & 1,349 & 64 & 1,364 & 10 \\
\hline
ast & val & 1,091 & 754 & 656 & 800 & 16 \\
\hline
ast & train & 163,466 & 151,164 & 86,392 & 149,505 & 2,385 \\
\hline
ast & test & 978 & 875 & 532 & 930 & 15 \\
\hline
az & val & 752 & 541 & 315 & 608 & 47 \\
\hline
az & train & 122,522 & 109,663 & 34,090 & 117,342 & 9,815 \\
\hline
az & test & 681 & 587 & 206 & 663 & 60 \\
\hline
azb & val & 1,141 & 887 & 308 & 991 & 180 \\
\hline
azb & train & 185,055 & 180,778 & 32,959 & 181,264 & 28,173 \\
\hline
azb & test & 1,080 & 1,035 & 213 & 1,062 & 179 \\
\hline
ba & val & 247 & 187 & 122 & 187 & 43 \\
\hline
ba & train & 39,870 & 37,115 & 16,325 & 37,453 & 7,996 \\
\hline
ba & test & 239 & 214 & 110 & 233 & 35 \\
\hline
bar & val & 199 & 143 & 115 & 141 & 24 \\
\hline
bar & train & 27,762 & 25,995 & 14,813 & 25,307 & 4,551 \\
\hline
bar & test & 177 & 155 & 93 & 168 & 34 \\
\hline
be & val & 896 & 679 & 362 & 763 & 112 \\
\hline
be & train & 147,532 & 129,774 & 51,508 & 138,881 & 19,117 \\
\hline
be & test & 845 & 736 & 308 & 812 & 112 \\
\hline
be-tarask & val & 385 & 285 & 149 & 325 & 38 \\
\hline
be-tarask & train & 64,362 & 56,142 & 22,373 & 60,059 & 7,673 \\
\hline
be-tarask & test & 336 & 303 & 138 & 318 & 46 \\
\hline
bg & val & 2,601 & 1,291 & 1,741 & 1,453 & 153 \\
\hline
bg & train & 282,629 & 246,330 & 137,138 & 257,584 & 29,052 \\
\hline
bg & test & 1,635 & 1,367 & 884 & 1,508 & 169 \\
\hline
bn & val & 643 & 411 & 415 & 467 & 45 \\
\hline
bn & train & 92,497 & 81,882 & 51,035 & 86,040 & 6,670 \\
\hline
bn & test & 590 & 493 & 343 & 563 & 48 \\
\hline
br & val & 435 & 298 & 315 & 299 & 25 \\
\hline
br & train & 65,252 & 60,404 & 41,839 & 55,855 & 4,228 \\
\hline
br & test & 410 & 363 & 279 & 363 & 21 \\
\hline
bs & val & 506 & 339 & 341 & 346 & 48 \\
\hline
bs & train & 72,527 & 66,194 & 39,723 & 65,378 & 8,072 \\
\hline
bs & test & 417 & 374 & 247 & 379 & 52 \\
\hline
ca & val & 4,310 & 2,980 & 1,913 & 3,455 & 267 \\
\hline
ca & train & 698,364 & 592,765 & 246,834 & 654,920 & 53,396 \\
\hline
ca & test & 3,818 & 3,185 & 1,410 & 3,637 & 252 \\
\hline
ce & val & 330 & 308 & 33 & 317 & 272 \\
\hline
ce & train & 54,669 & 54,043 & 2,983 & 54,122 & 47,050 \\
\hline
ce & test & 278 & 276 & 20 & 273 & 237 \\
\hline
ceb & val & 1,467 & 1,278 & 19 & 1,458 & 3 \\
\hline
ceb & train & 273,344 & 251,059 & 3,830 & 272,904 & 2,082 \\
\hline
ceb & test & 1,388 & 1,277 & 17 & 1,386 & 4 \\
\hline
ckb & val & 138 & 76 & 99 & 81 & 11 \\
\hline
ckb & train & 15,424 & 14,321 & 8,811 & 13,800 & 1,283 \\
\hline
ckb & test & 109 & 102 & 65 & 103 & 6 \\
\hline

\end{tabular}
\end{table*}

\begin{table*}
  \caption{Language Stats}
  \label{tab:wit-lang-stats-2}
  \begin{tabular}{|c|c|c|c|c|c|c|}
    \hline
    \textbf{Lang}
    & \textbf{Dataset}
    & \textbf{\# Examples}
    & \textbf{\# Uniq. Img}
    & \textbf{\# Ref}
    & \textbf{\# Attr}
    & \textbf{\# Cap}
    \\
  \hline

cs & val & 4,291 & 2,855 & 2,528 & 3,499 & 877 \\
\hline
cs & train & 652,034 & 557,020 & 332,616 & 613,059 & 147,478 \\
\hline
cs & test & 3,447 & 2,886 & 1,767 & 3,273 & 724 \\
\hline
cv & val & 118 & 89 & 60 & 87 & 18 \\
\hline
cv & train & 14,742 & 14,013 & 6,467 & 13,337 & 1,867 \\
\hline
cv & test & 103 & 96 & 51 & 94 & 11 \\
\hline
cy & val & 773 & 535 & 221 & 656 & 19 \\
\hline
cy & train & 122,238 & 106,448 & 21,917 & 117,650 & 2,966 \\
\hline
cy & test & 720 & 618 & 159 & 697 & 25 \\
\hline
da & val & 1,707 & 1,117 & 1,177 & 1,211 & 79 \\
\hline
da & train & 235,285 & 209,386 & 142,343 & 212,290 & 16,701 \\
\hline
da & test & 1,397 & 1,211 & 864 & 1,272 & 96 \\
\hline
de & val & 23,589 & 13,374 & 14,377 & 17,954 & 3,973 \\
\hline
de & train & 3,350,887 & 2,632,431 & 1,720,948 & 3,192,611 & 721,886 \\
\hline
de & test & 17,507 & 13,205 & 9,413 & 16,806 & 3,539 \\
\hline
el & val & 1,312 & 808 & 766 & 945 & 88 \\
\hline
el & train & 187,283 & 165,482 & 94,120 & 170,918 & 5,664 \\
\hline
el & test & 1,119 & 959 & 589 & 1,055 & 41 \\
\hline
en & val & 45,542 & 19,979 & 33,173 & 29,385 & 3,193 \\
\hline
en & train & 5,422,027 & 3,940,784 & 3,264,650 & 5,094,378 & 561,756 \\
\hline
en & test & 33,177 & 21,870 & 22,439 & 31,699 & 3,662 \\
\hline
eo & val & 1,771 & 1,247 & 1,103 & 1,343 & 232 \\
\hline
eo & train & 286,521 & 254,160 & 156,601 & 260,429 & 44,127 \\
\hline
eo & test & 1,589 & 1,339 & 885 & 1,484 & 265 \\
\hline
es & val & 13,390 & 7,378 & 9,644 & 8,552 & 221 \\
\hline
es & train & 1,741,015 & 1,433,075 & 1,092,875 & 1,530,281 & 37,151 \\
\hline
es & test & 9,876 & 7,658 & 6,461 & 9,001 & 198 \\
\hline
et & val & 1,372 & 771 & 1,009 & 865 & 49 \\
\hline
et & train & 169,304 & 150,267 & 109,891 & 150,847 & 9,761 \\
\hline
et & test & 966 & 835 & 666 & 872 & 50 \\
\hline
eu & val & 1,951 & 1,520 & 549 & 1,715 & 27 \\
\hline
eu & train & 321,311 & 294,256 & 63,662 & 310,440 & 4,914 \\
\hline
eu & test & 1,771 & 1,604 & 393 & 1,731 & 15 \\
\hline
fa & val & 3,162 & 2,215 & 1,305 & 2,562 & 389 \\
\hline
fa & train & 487,183 & 429,833 & 148,916 & 468,121 & 73,117 \\
\hline
fa & test & 2,839 & 2,447 & 962 & 2,758 & 437 \\
\hline
fi & val & 2,493 & 1,709 & 1,490 & 1,846 & 343 \\
\hline
fi & train & 377,569 & 333,999 & 188,244 & 348,713 & 64,404 \\
\hline
fi & test & 2,166 & 1,897 & 1,100 & 2,032 & 401 \\
\hline
fil & val & 273 & 193 & 149 & 221 & 21 \\
\hline
fil & train & 37,964 & 35,009 & 18,968 & 34,446 & 5,221 \\
\hline
fil & test & 353 & 242 & 207 & 269 & 33 \\
\hline
fr & val & 16,822 & 10,221 & 8,243 & 13,463 & 3,579 \\
\hline
fr & train & 2,561,756 & 2,018,324 & 976,576 & 2,444,258 & 658,134 \\
\hline
fr & test & 13,943 & 10,479 & 5,851 & 13,444 & 3,521 \\
\hline
fy & val & 353 & 218 & 248 & 258 & 17 \\
\hline
fy & train & 49,108 & 44,420 & 31,821 & 44,169 & 2,944 \\
\hline
fy & test & 328 & 292 & 227 & 308 & 23 \\
\hline
ga & val & 501 & 159 & 461 & 171 & 6 \\
\hline
ga & train & 33,094 & 31,344 & 25,471 & 28,807 & 1,404 \\
\hline
ga & test & 232 & 217 & 180 & 212 & 9 \\
\hline
gl & val & 1,244 & 743 & 817 & 827 & 36 \\
\hline
gl & train & 176,492 & 151,579 & 96,982 & 157,170 & 6,876 \\
\hline
gl & test & 1,032 & 873 & 579 & 953 & 46 \\
\hline
hi & val & 486 & 327 & 306 & 358 & 24 \\
\hline
hi & train & 69,397 & 61,813 & 38,280 & 62,967 & 4,929 \\
\hline
hi & test & 542 & 448 & 347 & 522 & 54 \\
\hline

\end{tabular}
\end{table*}

\begin{table*}
  \caption{Language Stats}
  \label{tab:wit-lang-stats-3}
  \begin{tabular}{|c|c|c|c|c|c|c|}
    \hline
    \textbf{Lang}
    & \textbf{Dataset}
    & \textbf{\# Examples}
    & \textbf{\# Uniq. Img}
    & \textbf{\# Ref}
    & \textbf{\# Attr}
    & \textbf{\# Cap}
    \\
  \hline

hr & val & 918 & 602 & 615 & 631 & 97 \\
\hline
hr & train & 139,111 & 118,919 & 79,178 & 123,083 & 18,866 \\
\hline
hr & test & 900 & 722 & 541 & 803 & 124 \\
\hline
hsb & val & 91 & 72 & 29 & 85 & 30 \\
\hline
hsb & train & 16,945 & 15,571 & 4,610 & 16,360 & 4,898 \\
\hline
hsb & test & 102 & 92 & 33 & 101 & 32 \\
\hline
ht & val & 142 & 120 & 35 & 119 & 69 \\
\hline
ht & train & 23,751 & 23,376 & 3,745 & 23,305 & 12,889 \\
\hline
ht & test & 148 & 144 & 21 & 147 & 90 \\
\hline
hu & val & 3,655 & 2,564 & 1,654 & 2,969 & 508 \\
\hline
hu & train & 563,738 & 495,429 & 209,013 & 519,752 & 86,780 \\
\hline
hu & test & 3,144 & 2,700 & 1,186 & 2,920 & 465 \\
\hline
hy & val & 1,713 & 1,235 & 670 & 1,379 & 37 \\
\hline
hy & train & 258,993 & 237,008 & 71,621 & 247,576 & 6,886 \\
\hline
hy & test & 1,454 & 1,289 & 419 & 1,401 & 52 \\
\hline
ia & val & 116 & 99 & 17 & 104 & 7 \\
\hline
ia & train & 16,666 & 16,250 & 2,075 & 16,112 & 698 \\
\hline
ia & test & 105 & 100 & 13 & 103 & 8 \\
\hline
id & val & 1,865 & 1,144 & 1,179 & 1,316 & 122 \\
\hline
id & train & 271,133 & 234,762 & 146,112 & 247,845 & 22,030 \\
\hline
id & test & 1,593 & 1,338 & 938 & 1,483 & 143 \\
\hline
io & val & 127 & 84 & 77 & 90 & 20 \\
\hline
io & train & 21,704 & 20,045 & 9,684 & 19,776 & 4,813 \\
\hline
io & test & 143 & 125 & 70 & 133 & 25 \\
\hline
is & val & 242 & 171 & 190 & 165 & 18 \\
\hline
is & train & 35,104 & 32,701 & 24,350 & 30,276 & 3,213 \\
\hline
is & test & 219 & 201 & 167 & 198 & 16 \\
\hline
it & val & 10,111 & 5,858 & 5,990 & 7,236 & 1,126 \\
\hline
it & train & 1,405,878 & 1,155,676 & 649,558 & 1,323,265 & 205,415 \\
\hline
it & test & 7,717 & 6,106 & 3,683 & 7,358 & 1,081 \\
\hline
iw & val & 2,775 & 1,610 & 1,636 & 2,168 & 152 \\
\hline
iw & train & 362,493 & 312,224 & 174,576 & 337,418 & 25,760 \\
\hline
iw & test & 2,091 & 1,740 & 1,052 & 1,949 & 137 \\
\hline
ja & val & 8,851 & 4,313 & 6,410 & 5,367 & 504 \\
\hline
ja & train & 1,084,871 & 874,699 & 624,537 & 1,003,353 & 92,839 \\
\hline
ja & test & 6,023 & 4,647 & 3,598 & 5,590 & 556 \\
\hline
jv & val & 130 & 99 & 83 & 97 & 6 \\
\hline
jv & train & 20,119 & 19,143 & 12,097 & 18,088 & 1,274 \\
\hline
jv & test & 122 & 116 & 73 & 119 & 8 \\
\hline
ka & val & 629 & 468 & 270 & 513 & 29 \\
\hline
ka & train & 104,856 & 94,129 & 36,443 & 97,517 & 8,248 \\
\hline
ka & test & 656 & 566 & 244 & 620 & 63 \\
\hline
kk & val & 325 & 245 & 150 & 266 & 69 \\
\hline
kk & train & 57,517 & 52,867 & 23,358 & 53,446 & 12,249 \\
\hline
kk & test & 336 & 297 & 158 & 320 & 63 \\
\hline
kn & val & 214 & 150 & 160 & 162 & 17 \\
\hline
kn & train & 30,406 & 28,060 & 21,687 & 27,315 & 1,566 \\
\hline
kn & test & 231 & 201 & 184 & 215 & 16 \\
\hline
ko & val & 2,065 & 1,386 & 1,138 & 1,573 & 183 \\
\hline
ko & train & 308,740 & 274,285 & 137,209 & 285,007 & 27,979 \\
\hline
ko & test & 1,802 & 1,586 & 881 & 1,674 & 165 \\
\hline
la & val & 1,209 & 680 & 1,027 & 658 & 96 \\
\hline
la & train & 142,674 & 134,587 & 108,875 & 124,441 & 18,944 \\
\hline
la & test & 742 & 690 & 569 & 674 & 101 \\
\hline
lah & val & 170 & 125 & 70 & 140 & 9 \\
\hline
lah & train & 30,825 & 27,456 & 9,590 & 29,383 & 1,634 \\
\hline
lah & test & 206 & 183 & 69 & 204 & 13 \\
\hline

\end{tabular}
\end{table*}

\begin{table*}
  \caption{Language Stats}
  \label{tab:wit-lang-stats-4}
  \begin{tabular}{|c|c|c|c|c|c|c|}
    \hline
    \textbf{Lang}
    & \textbf{Dataset}
    & \textbf{\# Examples}
    & \textbf{\# Uniq. Img}
    & \textbf{\# Ref}
    & \textbf{\# Attr}
    & \textbf{\# Cap}
    \\
  \hline

lb & val & 324 & 227 & 195 & 243 & 14 \\
\hline
lb & train & 49,401 & 43,998 & 22,042 & 46,565 & 3,278 \\
\hline
lb & test & 274 & 251 & 130 & 260 & 19 \\
\hline
lmo & val & 156 & 122 & 71 & 112 & 31 \\
\hline
lmo & train & 23,815 & 23,058 & 9,675 & 21,607 & 4,387 \\
\hline
lmo & test & 124 & 118 & 56 & 113 & 26 \\
\hline
lt & val & 967 & 723 & 491 & 774 & 77 \\
\hline
lt & train & 150,747 & 137,380 & 62,512 & 139,776 & 12,861 \\
\hline
lt & test & 817 & 728 & 351 & 772 & 71 \\
\hline
lv & val & 609 & 451 & 311 & 501 & 139 \\
\hline
lv & train & 95,165 & 83,567 & 43,276 & 88,648 & 25,237 \\
\hline
lv & test & 573 & 489 & 278 & 522 & 172 \\
\hline
mg & val & 115 & 96 & 41 & 81 & 45 \\
\hline
mg & train & 21,397 & 21,200 & 5,077 & 17,919 & 57 \\
\hline
mg & test & 143 & 142 & 23 & 127 & 15,277 \\
\hline
mk & val & 723 & 485 & 431 & 492 & 75 \\
\hline
mk & train & 113,482 & 97,839 & 59,047 & 103,236 & 25 \\
\hline
mk & test & 676 & 591 & 366 & 626 & 6,041 \\
\hline
ml & val & 564 & 371 & 322 & 424 & 54 \\
\hline
ml & train & 90,149 & 80,683 & 42,723 & 84,472 & 10 \\
\hline
ml & test & 570 & 494 & 292 & 546 & 1,990 \\
\hline
mn & val & 128 & 104 & 73 & 100 & 8 \\
\hline
mn & train & 19,695 & 18,715 & 8,167 & 18,307 & 14 \\
\hline
mn & test & 109 & 102 & 48 & 105 & 1,621 \\
\hline
mr & val & 169 & 131 & 93 & 141 & 6 \\
\hline
mr & train & 24,262 & 22,343 & 12,210 & 22,664 & 63 \\
\hline
mr & test & 165 & 148 & 85 & 162 & 11,330 \\
\hline
ms & val & 728 & 492 & 444 & 562 & 88 \\
\hline
ms & train & 113,190 & 103,204 & 57,894 & 103,981 & 3 \\
\hline
ms & test & 687 & 620 & 389 & 630 & 1,537 \\
\hline
my & val & 107 & 67 & 71 & 79 & 11 \\
\hline
my & train & 17,259 & 16,083 & 10,015 & 15,961 & 739 \\
\hline
my & test & 104 & 97 & 59 & 97 & 131,801 \\
\hline
nan & val & 1,387 & 831 & 603 & 840 & 631 \\
\hline
nan & train & 161,287 & 159,564 & 19,725 & 152,725 & 3 \\
\hline
nan & test & 809 & 792 & 126 & 765 & 888 \\
\hline
nds & val & 144 & 106 & 127 & 99 & 11 \\
\hline
nds & train & 19,024 & 17,954 & 15,735 & 15,998 & 7 \\
\hline
nds & test & 118 & 105 & 97 & 113 & 1,554 \\
\hline
ne & val & 128 & 100 & 66 & 108 & 9 \\
\hline
ne & train & 21,415 & 19,197 & 9,177 & 20,430 & 1,993 \\
\hline
ne & test & 150 & 134 & 72 & 150 & 360,360 \\
\hline
nl & val & 7,888 & 5,014 & 3,434 & 6,524 & 1,825 \\
\hline
nl & train & 1,225,211 & 969,736 & 405,788 & 1,169,330 & 87 \\
\hline
nl & test & 6,511 & 4,975 & 2,300 & 6,244 & 15,892 \\
\hline
nn & val & 796 & 535 & 498 & 589 & 101 \\
\hline
nn & train & 117,036 & 106,463 & 62,556 & 105,519 & 365 \\
\hline
nn & test & 757 & 639 & 413 & 700 & 64,330 \\
\hline
no & val & 2,971 & 1,940 & 1,759 & 2,316 & 452 \\
\hline
no & train & 430,419 & 362,192 & 219,405 & 397,910 & 2 \\
\hline
no & test & 2,662 & 2,072 & 1,510 & 2,496 & 6 \\
\hline
nv & val & 118 & 106 & 20 & 117 & 2,265 \\
\hline
nv & train & 20,691 & 18,386 & 3,300 & 20,479 & 13 \\
\hline
nv & test & 92 & 83 & 21 & 89 & 5 \\
\hline
oc & val & 347 & 251 & 215 & 249 & 1,354 \\
\hline
oc & train & 62,915 & 59,145 & 32,093 & 56,856 & 22 \\
\hline
oc & test & 363 & 326 & 209 & 325 & 2,790 \\
\hline

\end{tabular}
\end{table*}

\begin{table*}
  \caption{Language Stats}
  \label{tab:wit-lang-stats-5}
  \begin{tabular}{|c|c|c|c|c|c|c|}
    \hline
    \textbf{Lang}
    & \textbf{Dataset}
    & \textbf{\# Examples}
    & \textbf{\# Uniq. Img}
    & \textbf{\# Ref}
    & \textbf{\# Attr}
    & \textbf{\# Cap}
    \\
  \hline

pa & val & 193 & 127 & 126 & 136 & 505,890 \\
\hline
pa & train & 27,122 & 25,126 & 15,053 & 25,113 & 2,642 \\
\hline
pa & test & 211 & 184 & 120 & 206 & 416 \\
\hline
pl & val & 7,036 & 4,773 & 3,308 & 5,783 & 73,480 \\
\hline
pl & train & 1,114,489 & 938,346 & 429,129 & 1,054,164 & 396 \\
\hline
pl & test & 5,747 & 4,798 & 2,235 & 5,507 & 16 \\
\hline
pt & val & 6,515 & 3,354 & 3,920 & 4,134 & 3,093 \\
\hline
pt & train & 804,395 & 672,754 & 332,753 & 754,910 & 26 \\
\hline
pt & test & 4,592 & 3,684 & 2,017 & 4,372 & 544 \\
\hline
qu & val & 117 & 75 & 61 & 94 & 102,809 \\
\hline
qu & train & 18,900 & 15,381 & 8,596 & 17,116 & 540 \\
\hline
qu & test & 134 & 103 & 65 & 128 & 797 \\
\hline
ro & val & 2,294 & 1,636 & 924 & 1,939 & 141,748 \\
\hline
ro & train & 375,263 & 324,990 & 116,427 & 357,865 & 764 \\
\hline
ro & test & 2,095 & 1,772 & 719 & 2,013 & 43 \\
\hline
ru & val & 9,808 & 6,225 & 4,445 & 8,067 & 8,537 \\
\hline
ru & train & 1,534,893 & 1,240,599 & 569,666 & 1,467,699 & 54 \\
\hline
ru & test & 8,780 & 6,792 & 3,610 & 8,485 & 7 \\
\hline
sco & val & 329 & 240 & 208 & 246 & 1,073 \\
\hline
sco & train & 47,619 & 45,243 & 28,392 & 42,170 & 12 \\
\hline
sco & test & 304 & 277 & 197 & 275 & 330 \\
\hline
si & val & 95 & 66 & 68 & 63 & 66,022 \\
\hline
si & train & 14,903 & 13,392 & 9,344 & 13,250 & 326 \\
\hline
si & test & 124 & 108 & 76 & 115 & 131 \\
\hline
sk & val & 1,433 & 1,111 & 520 & 1,234 & 27,002 \\
\hline
sk & train & 229,755 & 209,237 & 62,766 & 218,817 & 167 \\
\hline
sk & test & 1,233 & 1,087 & 385 & 1,180 & 59 \\
\hline
sl & val & 1,055 & 673 & 638 & 746 & 11,113 \\
\hline
sl & train & 146,551 & 130,395 & 71,919 & 134,643 & 51 \\
\hline
sl & test & 839 & 735 & 429 & 793 & 100 \\
\hline
sq & val & 355 & 262 & 214 & 263 & 18,810 \\
\hline
sq & train & 48,942 & 44,433 & 22,455 & 45,258 & 103 \\
\hline
sq & test & 273 & 243 & 133 & 252 & 58 \\
\hline
sr & val & 2,649 & 1,538 & 1,617 & 1,719 & 11,953 \\
\hline
sr & train & 358,505 & 314,625 & 166,598 & 324,088 & 81 \\
\hline
sr & test & 2,000 & 1,720 & 1,010 & 1,827 & 450 \\
\hline
sr-Latn & val & 1,007 & 723 & 563 & 707 & 87,196 \\
\hline
sr-Latn & train & 165,373 & 150,072 & 75,782 & 143,264 & 520 \\
\hline
sr-Latn & test & 957 & 840 & 463 & 847 & 11 \\
\hline
sv & val & 6,330 & 4,058 & 3,252 & 4,817 & 2,259 \\
\hline
sv & train & 918,732 & 791,029 & 363,165 & 877,774 & 10 \\
\hline
sv & test & 4,914 & 4,171 & 2,130 & 4,732 & 33 \\
\hline
sw & val & 201 & 136 & 153 & 131 & 5,963 \\
\hline
sw & train & 28,993 & 27,225 & 22,141 & 24,424 & 31 \\
\hline
sw & test & 164 & 151 & 131 & 145 & 17 \\
\hline
ta & val & 608 & 362 & 410 & 424 & 2,501 \\
\hline
ta & train & 85,453 & 75,394 & 48,154 & 77,670 & 17 \\
\hline
ta & test & 526 & 446 & 318 & 496 & 11 \\
\hline
te & val & 279 & 160 & 190 & 189 & 1,823 \\
\hline
te & train & 37,589 & 33,063 & 23,336 & 34,776 & 14 \\
\hline
te & test & 267 & 223 & 168 & 255 & 37 \\
\hline
tg & val & 155 & 125 & 57 & 132 & 8,480 \\
\hline
tg & train & 22,594 & 21,557 & 6,660 & 21,883 & 58 \\
\hline
tg & test & 143 & 137 & 38 & 142 & 142 \\
\hline
th & val & 793 & 544 & 437 & 579 & 25,437 \\
\hline
th & train & 120,592 & 105,618 & 54,853 & 110,325 & 151 \\
\hline
th & test & 750 & 633 & 366 & 697 & 12 \\
\hline

\end{tabular}
\end{table*}

\begin{table*}
  \caption{Language Stats}
  \label{tab:wit-lang-stats-6}
  \begin{tabular}{|c|c|c|c|c|c|c|}
    \hline
    \textbf{Lang}
    & \textbf{Dataset}
    & \textbf{\# Examples}
    & \textbf{\# Uniq. Img}
    & \textbf{\# Ref}
    & \textbf{\# Attr}
    & \textbf{\# Cap}
    \\
  \hline

tr & val & 1,840 & 1,265 & 1,023 & 1,413 & 2,340 \\
\hline
tr & train & 270,292 & 237,703 & 126,042 & 250,380 & 10 \\
\hline
tr & test & 1,591 & 1,366 & 762 & 1,509 & 430 \\
\hline
tt & val & 274 & 225 & 85 & 242 & 78,021 \\
\hline
tt & train & 44,622 & 42,005 & 9,852 & 42,849 & 464 \\
\hline
tt & test & 285 & 268 & 74 & 274 & 173 \\
\hline
uk & val & 5,990 & 4,135 & 2,316 & 4,983 & 27,968 \\
\hline
uk & train & 924,811 & 787,944 & 279,548 & 885,861 & 176 \\
\hline
uk & test & 4,985 & 4,177 & 1,503 & 4,837 & 165 \\
\hline
ur & val & 567 & 413 & 291 & 475 & 27,394 \\
\hline
ur & train & 91,264 & 82,093 & 36,659 & 86,497 & 142 \\
\hline
ur & test & 580 & 489 & 246 & 558 & 251 \\
\hline
uz & val & 272 & 243 & 64 & 249 & 42,060 \\
\hline
uz & train & 42,808 & 42,084 & 7,722 & 41,971 & 205 \\
\hline
uz & test & 239 & 238 & 59 & 239 & 310 \\
\hline
vec & val & 290 & 268 & 25 & 278 & 61,086 \\
\hline
vec & train & 48,684 & 48,262 & 3,728 & 47,735 & 330 \\
\hline
vec & test & 265 & 264 & 40 & 256 & 3 \\
\hline
vi & val & 3,351 & 2,321 & 1,467 & 2,719 & 619 \\
\hline
vi & train & 515,937 & 453,326 & 177,565 & 491,288 & 4 \\
\hline
vi & test & 2,696 & 2,311 & 1,051 & 2,593 & 18 \\
\hline
vo & val & 180 & 157 & 167 & 86 & 4,636 \\
\hline
vo & train & 30,564 & 29,293 & 27,485 & 16,248 & 17 \\
\hline
vo & test & 134 & 127 & 116 & 83 & 2 \\
\hline
war & val & 443 & 354 & 72 & 400 & 1,213 \\
\hline
war & train & 75,167 & 68,979 & 9,022 & 71,917 & 13 \\
\hline
war & test & 334 & 312 & 45 & 316 & 11 \\
\hline
xmf & val & 86 & 77 & 33 & 68 & 1,807 \\
\hline
xmf & train & 15,221 & 14,405 & 5,562 & 13,801 & 11 \\
\hline
xmf & test & 92 & 89 & 27 & 89 & 292 \\
\hline
yue & val & 333 & 260 & 243 & 255 & 58,212 \\
\hline
yue & train & 52,494 & 49,464 & 34,196 & 46,744 & 357 \\
\hline
yue & test & 309 & 293 & 215 & 286 & 291 \\
\hline
zh & val & 5,493 & 3,504 & 3,164 & 4,306 & 58,289 \\
\hline
zh & train & 834,324 & 691,440 & 411,939 & 784,829 & 358 \\
\hline
zh & test & 4,947 & 3,905 & 2,733 & 4,716 &  \\
\hline
zh-TW & val & 5,493 & 3,508 & 3,157 & 4,312 &  \\
\hline
zh-TW & train & 834,941 & 691,337 & 412,317 & 785,378 &  \\
\hline
zh-TW & test & 4,955 & 3,903 & 2,736 & 4,721 &  \\
\hline

\end{tabular}
\end{table*}

\end{document}